# Multi-hop Inference for Sentence-level TextGraphs: How Challenging is Meaningfully Combining Information for Science Question Answering?

Peter A. Jansen
School of Information, University of Arizona, Tucson, AZ
pajansen@email.arizona.edu

## Abstract

Question Answering for complex questions is often modelled as a graph construction or traversal task, where a solver must build or traverse a graph of facts that answer and explain a given question. This "multi-hop" inference has been shown to be extremely challenging, with few models able to aggregate more than two facts before being overwhelmed by "semantic drift", or the tendency for long chains of facts to quickly drift off topic. This is a major barrier to current inference models, as even elementary science questions require an average of 4 to 6 facts to answer and explain. In this work we empirically characterize the difficulty of building or traversing a graph of sentences connected by lexical overlap, by evaluating chance sentence aggregation quality through 9,784 manually-annotated judgements across knowledge graphs built from three free-text corpora (including study guides and Simple Wikipedia). We demonstrate semantic drift tends to be high and aggregation quality low, at between 0.04% and 3%, and highlight scenarios that maximize the likelihood of meaningfully combining information.

## 1 Introduction

Question answering (QA) is a task where models must find answers to natural language questions, either by retrieving these answers from a corpus, or inferring them by some inference process. Retrieval methods model QA as an answer sentence selection task, where a solver must find a sentence or short continuous passage of text in a corpus that answers the question (Moschitti et al., 2007; Severyn and Moschitti, 2012, inter alia). These methods often fall short for questions requiring complex inference, such as those in the science domain, where nearly 80% of even $4^{th}$ grade science exam questions require some form of causal, model-based, or otherwise com-plex inference to answer and explain (Clark et al., 2013; Jansen et al., 2016), and a single continuous passage of text rarely describes the reasoning required to move from question to correct answer. In these cases, *multiple sentences*, often from different parts of a text, different documents, or different knowledge bases must be aggregated together to build a complete answer and explanation.

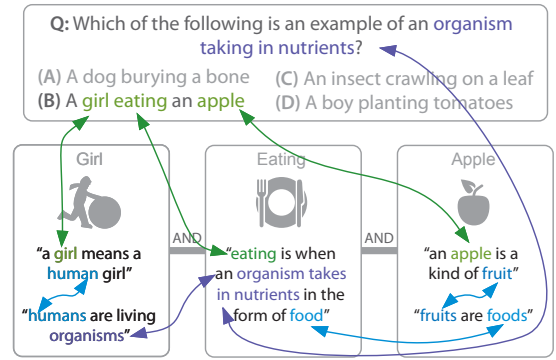

Figure 1: An example multiple choice $4^{th}$ grade science question from the NY Regents exam, and a graph of 5 sentences that answer and explain the answer to this question. Edges represent lexical overlap.

Aggregating knowledge to support inference and complex question answering is often framed as a graph construction or traversal problem (e.g. Khashabi et al., 2016), where the solver must find paths that link sentences that contain question terms with sentences that contain answer terms through some number of intermediate sentences (see Figure 1). In these knowledge graphs, nodes represent facts or single sentences, and edges between nodes represent some signal that the facts are interrelated, such as having lexical overlap.

Information aggregation or "multi-hop" graph traversal has been shown to be extremely challenging, with QA solvers generally showing only modest performance benefits when aggregating information, and diminishing returns as the amount of aggregation increases. In the elementary sci-ence domain, current estimates suggest that an average of 4 to 6 sentences are required to answer and explain a given question (Jansen et al., 2016, 2018), while recent QA solvers generally struggle to meaningfully aggregate more than two free-text sentences (Jansen et al., 2017), even when using alternate representations including semi-structured tables (Khashabi et al., 2016) or graphs of words or syntactic dependencies traversed using monolingual alignment or PageRank variants in open-domain QA (Fried et al., 2015). Fried et al. (2015) suggest these performance limitations are due to *"semantic drift"*, where as the number of sentences being aggregated increases, so do the chances of making a misstep in the aggregation – for example, aggregating a sentence about *seed funding for a company* when making an inference about the *stages of plant growth*. This appears to occur across a variety of solvers, representations, and methods for aggregation, and is leading to both the development of datasets specifically designed for multi-hop QA (Jansen et al., 2016, 2018; Welbl et al., 2017), as well as methods of controlling for semantic drift in knowledge graphs constructed from (for example) OpenIE triples using either support graphs (Khot et al., 2017) or drift-sensitive random walks (Kwon et al., 2018).

In an effort to better understand the challenges of inference and explanation construction for QA, here we characterize the difficultly of the information aggregation task in the context of science exams. The contributions of this work are:

1. We provide the first empirical characterization of the difficulty of information aggregation by manually evaluating sentence aggregation quality using 9,784 annotated judgements across 14 representative exam questions, highlighting specific patterns of lexical overlap between question, answer, and candidate sentence that maximize the chances of successful aggregation.

2. We evaluate aggregation difficulty across three knowledge resources, and empirically demonstrate that while moving to open domain resources increases knowledge coverage, it also increases the difficulty of the information aggregation task by more than an order of magnitude.

3. We evaluate aggregating up to three sentences that connect terms in the question to terms in the answer, and show that this suffers both from sparsity (even on Wikipedia-scale corpora), as well as a very low probability of producing meaningful aggregations (0.04% to 3%) through lexical overlap alone.

## 2 Approach

**Questions:** Due to the magnitude of manual annotation, we drew 14 representative questions annotated as likely requiring inference[1] from the 432 training questions in the AI2 Open Elementary Science Questions set[2], originally drawn from standardized science exams in 12 US states. Questions span 14 common curriculum topics, including changes of state, planetary motion, environmental adaptations, the life cycle, inherited traits, magnetism, and measurement. For context, to date, the best-performing systems report answering just under 60% of elementary science questions correctly (Jauhar et al., 2016).

### 2.1 Corpora

We generate and evaluate three separate graphs constructed from three independent corpora:

**Science Explanations Corpus:** An explanation corpus of 1,364 sentences from Jansen et al. (2016) designed to construct high-quality explanations for the AI2 question set through aggregation.

**Study Guide Corpus:** An in-domain corpus of 2,503 sentences from two study guides for the New York and Virginia standardized exams.

**Simple Wikipedia:** A large open-domain corpus of 848,920 sentences retrieved from Simple Wikipedia and included in the AristoMini corpus[2].

### 2.2 Methods

Here we simulate the graph-based inference process by creating short chains of sentences interconnected based on shared words between those sentences. Specifically, two sentences are said to be connected if they share at least one content lemma *(noun, verb, or adjective)* in common. Sentences with the same lemma but different parts of speech are not connected (e.g. a sentence containing *plant_VB* is not connected to a sentence containing *plant_NN*). Lemmatization and part-of-speech tagging are provided by the Stanford CoreNLP toolkit (Manning et al., 2014).

---

[1]Our results did not substantially change when data from only half the questions were used, suggesting the aggregate statistics from the 9,784 manual judgements are stable.

[2]http://allenai.org/data.html

*Q: What is the main purpose of the flowers of a peach tree?*
**A: to attract bees for pollination.**

*Example Ratings:*
**High:** The flower helps the plant reproduce because it contains the pollen and eggs.
**Possible:** Seeds grow in the center of a flower and continue to develop there after the petals fall off the plant.
**Topical/Unlikely:** There are four major parts of a plant: roots, stem, leaves, and flower.
**Offtopic:** The average life span of a worker bee is 1 year.

Table 1: Example ratings on a 4-point rating scale describing the perceived utility of each sentence towards an explanation for why the answer is correct (high, possible, topical, offtopic). Sentences are from the Study Guide corpus, and each have lexical overlap with the question and/or answer.

For a given question, sentences in one corpus are identified that have lexical overlap with either the question terms, answer terms, or both question and answer terms. We then manually rate the relevance of each sentence on a 4-point scale using the following criterion: *"What is the likelihood that this knowledge would contribute to an explanation for why the answer is correct?"*. Example ratings are included in Table 1.

### 2.3 Connectivity Characterization

Here, we denote the question text as $Q$, the correct answer text as $A$, and a sentence from the corpus with overlapping terms as $S_x$, where $x$ is either $Q$ or $A$. We characterize the utility of sentences towards building an explanation in five scenarios:

*Direct lexical overlap:*

1. $Q \leftrightarrow S_Q$: Sentences that have lexical overlap with the question.
2. $S_A \leftrightarrow A$: Sentences that have lexical overlap with the answer.
3. $Q \leftrightarrow S_{QA} \leftrightarrow A$: Sentences that have lexical overlap with both question and answer.

*Indirect (aggregating) overlap:*

4. $Q \leftrightarrow S_Q \leftrightarrow S_A \leftrightarrow A$: Aggregating two sentences that individually have lexical overlap with the question or answer, and that also have lexical overlap with each other.
5. $Q \leftrightarrow S_Q \leftrightarrow S_O \leftrightarrow S_A \leftrightarrow A$: Aggregating three sentences: two sentences that individually have lexical overlap with the question or answer, and that are connected by a third sentence $S_O$ that has lexical overlap with both $S_Q$ and $S_A$, but not with $Q$ or $A$.

## 3 Results and Discussion

**What proportion of sentences with direct lexical overlap to the question and answer contain highly relevant information?** The results of the direct characterization are shown in Table 2. The overall proportion of corpus sentences containing relevant information to the question are low, with 5.5% of sentences rated as highly useful in the explanation corpus, 1.7% in the Study Guide corpus, and only 0.1% in the large Simple Wikipedia corpus. Sentence utility increases as the lexical overlap (number of terms matched) increases. Similarly, sentences with terms from the answer are *3 to 5* times more likely to be highly relevant than sentences with question terms. Sentences that overlap on both question and answer terms have a substantially increased probability of being rated highly relevant compared to sentences with a single question or answer term (e.g. 21.4% vs 1.7% and 5.2%, respectively, for the Study Guide corpus), but are sparse, occurring an average of approximately once per question.

**When aggregating two sentences, what proportion will contain highly relevant information?** The probability of aggregating two sentences that individually lexically overlap with the question or answer, and also lexically overlap with each other, $Q \leftrightarrow S_Q \leftrightarrow S_A \leftrightarrow A$, is shown in Table 3. The likelihood of aggregating two sentences from the Study Guide corpus that were both highly rated and that lexically overlap by at least one term is 3.0%, and when expanding this to allow for aggregating sentences with high or possible ratings *(bolded square)*, this likelihood increases to 6.6%. For the Simple Wikipedia corpus these probabilities are one to two orders of magnitude lower, at 0.04% and 0.3%, respectively.

When restricting 2-sentence aggregations to cases of moderate lexical overlap, where $S_Q \leftrightarrow S_A$ overlap by *2 or more lemmas* not found in the question or answer, quality improves substantially

---

[3]The scale of the Simple Wikipedia corpus makes manual evaluation intractable. Here we subsample to rate 50 sentences with each pattern of lexical overlap, and limit our analysis to 7 questions. For example, for the question in in Table 1, we rate 50 sentences that have lexical overlap only with the word *flowers_NN*, another 50 that overlap with *flowers_NN* and *purpose_NN*, and so on. In practice, due to the relative sparsity of multiword matches, *we evaluate nearly all cases where the lexical overlap consists of two or more words*, and the subsampling only affects estimates of single overlapping lemma matches for this corpus *(leftmost columns in table)*.

|  | 1 overlapping lemma | | | 2 overlapping lemmas | | | 3+ overlapping lemmas | | |
|---|---|---|---|---|---|---|---|---|---|
|  | $S_Q$ | $S_A$ | $S_{QA}$ | $S_Q$ | $S_A$ | $S_{QA}$ | $S_Q$ | $S_A$ | $S_{QA}$ |
| *Explanation Corpus (1,364 sentences)* | | | | | | | | | |
| *Highly likely* | 5.5% | 18.4% | - | 18.2% | 66.7% | 65.6% | 40.0% | - | 100% |
| *Possible* | 4.8% | 8.5% | - | 22.7% | 6.7% | 9.4% | 40.0% | - | 0% |
| *Topical* | 14.1% | 26.9% | - | 18.2% | 0% | 18.8% | 0% | - | 0% |
| *Off topic* | 75.5% | 46.2% | - | 40.9% | 26.7% | 6.3% | 20.0% | - | 0% |
| *N (Samples)* | 992 | 223 | - | 44 | 15 | 32 | 5 | 0 | 7 |
| *Study Guide Corpus (2,503 sentences)* | | | | | | | | | |
| *Highly likely* | 1.7% | 5.2% | - | 6.6% | 55.6% | 21.4% | 27.8% | - | 58.8% |
| *Possible* | 2.1% | 6.3% | - | 12.4% | 5.6% | 12.5% | 27.8% | - | 17.6% |
| *Topical* | 6.7% | 9.6% | - | 24.1% | 16.7% | 12.5% | 11.1% | - | 5.9% |
| *Off topic* | 89.6% | 79.0% | - | 56.9% | 22.2% | 53.6% | 33.3% | - | 17.6% |
| *N (Samples)* | 2133 | 480 | - | 137 | 18 | 56 | 18 | 0 | 17 |
| *Simple Wikipedia Corpus (848,920 sentences, subsampled[3])* | | | | | | | | | |
| *Highly likely* | 0.1% | 0.5% | - | 0.4% | 11.2% | 1.7% | 0.6% | 50.0% | 16.1% |
| *Possible* | 0.2% | 1.1% | - | 1.8% | 6.7% | 3.3% | 2.5% | 50.0% | 19.4% |
| *Topical* | 0.8% | 3.4% | - | 3.4% | 17.9% | 8.2% | 5.0% | 0.0% | 14.5% |
| *Off topic* | 98.9% | 95.0% | - | 94.4% | 64.2% | 86.8% | 91.9% | 0.0% | 50.0% |
| *N (Samples)* | 2102 | 880 | - | 1399 | 134 | 599 | 161 | 2 | 62 |

Table 2: Observed frequencies for sentences with given utility ratings for the three categories of direct (lexical overlap) connections: $Q \leftrightarrow S_Q$, $S_A \leftrightarrow A$, and $Q \leftrightarrow S_{QA} \leftrightarrow A$, and various degrees of lexical overlap.

*Explanation Corpus (1,979 samples)*

|  |  | $S_A$ Rating | | | |
|---|---|---|---|---|---|
| | | Highly | Possible | Topical | OffTopic |
| $S_Q$ Rating | *Highly* | **13.4%** | **3.8%** | 3.4% | 1.1% |
| | *Possible* | **3.4%** | **0.4%** | 1.1% | 0.6% |
| | *Topical* | 7.5% | 1.3% | 8.0% | 4.5% |
| | *Offtopic* | 16.3% | 11.0% | 9.1% | 14.9% |

*Study Guide Corpus (8,096 samples)*

|  |  | $S_A$ Rating | | | |
|---|---|---|---|---|---|
| | | Highly | Possible | Topical | OffTopic |
| $S_Q$ Rating | *Highly* | **3.0%** | **0.7%** | 0.7% | 1.3% |
| | *Possible* | **2.1%** | **0.8%** | 0.8% | 1.2% |
| | *Topical* | 3.8% | 2.0% | 2.7% | 4.4% |
| | *Offtopic* | 17.0% | 6.4% | 10.9% | 42.2% |

*Simple Wikipedia Corpus (23,750 samples)*

|  |  | $S_A$ Rating | | | |
|---|---|---|---|---|---|
| | | Highly | Possible | Topical | OffTopic |
| $S_Q$ Rating | *Highly* | **0.04%** | **0.04%** | 0.06% | 0.4% |
| | *Possible* | **0.1%** | **0.1%** | 0.3% | 1.7% |
| | *Topical* | 0.1% | 0.02% | 0.2% | 2.1% |
| | *Offtopic* | 2.4% | 2.0% | 7.2% | 82.9% |

Table 3: Observed frequencies for aggregating two sentences together with specific utility ratings in the $Q \leftrightarrow S_Q \leftrightarrow S_A \leftrightarrow A$ condition across each corpus. Here, one sentence in the pair has overlapping terms in the question, the other sentence has overlapping terms in the answer, and both sentences lexically overlap with each other on *one or more* terms that are not found in either the question or answer. Axes represent the individual (nonaggregated) ratings of each sentence (Q or A). The bolded square represents the proportion of lexically connected sentence pairs where utility ratings for both sentences are either *high* or *possible*. Detailed versions of these tables can be found in the *Appendix*.

on the Study Guide corpus, with 12.5% of these aggregates containing sentences both rated *highly relevant* (N=1,262), or an average of 11 per question. The pattern is similar for the Explanation and SimpleWiki corpora, but scaled up by a factor of 2-4, and down by a factor of 10-40, respectively.[4]

**When aggregating three sentences, what proportion of intermediate sentences are highly relevant?** To characterize the number of possible 3-sentence aggregations of the form $Q \leftrightarrow S_Q \leftrightarrow S_O \leftrightarrow S_A \leftrightarrow A$, with each sentence rated as having a *highly relevant* or *possible* utility for explanations, we retrieved all intermediate sentences $S_O$ in the corpus such that (a) $S_O$ contains overlapping lemmas with both $S_Q$ and $S_A$ that are not found in the question or answer, and (b) both $S_Q$ and $S_A$ have ratings of either *highly relevant* or *possible*. The overall number of intermediate sentences meeting this criterion was small (17 for the Study Guide corpus across all 14 questions, and 251 for the Simple Wikipedia corpus). We manually rated these intermediate sentences, finding a small proportion had favourable utility ratings, with 1.5% receiving ratings of *highly relevant* and 2% receiving *possible*. This suggests that both sparsity and drift make aggregating three sentences highly unlikely, even in large million-sentence-scale corpora such as Simple Wikipedia.

---
[4]Due to space limitations, this table is not shown.

**Overall, what is chance performance for combining information to generate real explanations?** Previous work suggests that real explanations for elementary science questions require aggregating an average of 4 to 6 separate facts to answer and explain (Jansen et al., 2016, 2018), with this value ranging between 1 fact to more than a dozen facts per question, depending on the amount of question-specific knowledge and world knowledge required. Extrapolating from our empirical analysis[5] suggests that the chance of generating a 4-fact aggregation of the form $Q \leftrightarrow S_Q \leftrightarrow S_O \leftrightarrow S_O \leftrightarrow S_A \leftrightarrow A$ is likely to be extremely improbable, at approximately 1 in 187,000 for the Study Guide corpus, and 1 in 17 million with the Simple Wikipedia corpus, in the case of sentences having a single overlapping lemma. Where $S_Q$ and $S_A$ share two overlapping lemmas with the question, this increases to approximately 1 in 7,000 for the Study Guide corpus, and 1 in 207,000 for the Simple Wikipedia corpus, but is still improbable.

**Building graphs based solely on lexical overlap captures only a fraction of the possible meaningful connections between knowledge in a corpus. How might this limitation affect this empirical analysis?** Lexical overlap is a common method of building knowledge graphs for QA (e.g. Khashabi et al., 2016; Jansen et al., 2017), as two sentences having the same words has been regarded as a strong signal that they may contain mutually beneficial content for the inference task. While other methods of connection, such as WordNet synsets to capture synonymy or word embeddings to capture associative relations, are likely to increase the recall of sentences in a corpus relevant to a given question, we hypothesize that lexical overlap – as poorly as we have shown it performs empirically – is likely a higher precision method of creating meaningful connections than these other connection methods. In this way we propose lexical overlap can be viewed as a baseline for other knowledge graph connection methodologies to be evaluated against.

**Evaluating the proportion of meaningful connections in graphs built from specific knowledge resources provides only a partial understanding of the challenges of information aggregation, because it doesn't capture how well specific inference methods may perform on a given knowledge graph.** A central limitation of this empirical evaluation is that it evaluates the probability of meaningfully assembling knowledge in three specific knowledge resources, rather than the empirical performance of specific inference algorithms on assembling knowledge towards the QA and explanation construction task with these specific resources. Combining information to form inferences is one of the central challenges in contemporary question answering, and few models appear able to consistently aggregate more than two facts in support of this inference task. While a variety of different methods of information aggregation have been proposed, our ultimate evaluation metric for many of these models has been the overall proportion of questions answered correctly, rather than a targeted evaluation of the information aggregation mechanism. Methods such as evaluating inference performance as the number of aggregation steps increases (e.g. Fried et al., 2015; Jansen et al., 2017) begin to provide insight on the efficacy of specific methods of information aggregation, but these methods must be paired with a knowledge graph with known connectivity properties to provide a detailed characterization of the performance of specific aggregation methods on the information aggregation task.

## 4 Conclusion

We empirically demonstrate that aggregating multiple sentences together to support inference for QA is extremely challenging. For the in-domain study guide corpus, only 3% of 2-sentence $Q \leftrightarrow S_Q \leftrightarrow S_A \leftrightarrow A$ aggregations were rated as highly useful, while this falls to 0.04% for the open domain corpus. In spite of the size of Simple Wikipedia, 3-sentence aggregations are sparse, and substantially reduce the chance of meaningfully aggregating sentences to the point of improbability. Taken together, our analysis suggests the ability to generate inferences incorporating *4 to 6 facts* required for the average question is unlikely without high-precision means of concept matching beyond lexical overlap, and methods of controlling for drift, or reducing drift through pairing with close-domain corpora. Our ratings for the open Explanation and Simple Wikipedia corpora are available at http://cognitiveai.org/explanationbank/.

---

[5] This extrapolation uses the empirically derived probability of a meaningful $S_O$ transition to be 3.5% (1.5% *highly* + 2.0% *possible*). Similarly, probabilities for $S_Q$ and $S_A$ add both *highly* and *possible* transition probabilities from Table 2.

## 5 Appendix

A natural expansion of the analysis of aggregation quality of 2-sentence $Q \leftrightarrow S_Q \leftrightarrow S_A \leftrightarrow A$ aggregations found in Table 3 is considering cases where $S_Q$ and $S_A$ share *more than 1 word* not found in either the question or the answer (*"other"* words), suggesting the sentences may have more in common, and have a higher chance of being meaningfully aggregated. Aggregation quality for each corpus broken down by cases where $S_Q$ and $S_A$ share *at least 1, 2, 3, or 4 words* can be found in Tables 4, 5, and 6 for the COLING Explanation, Study Guide, and Simple Wikipedia corpora, respectively. Figure 2 summarizes these tables by showing the overall proportion of "good" 2-sentence $Q \leftrightarrow S_Q \leftrightarrow S_A \leftrightarrow A$ aggregations broken down by corpus. While increasing the minimum overlap threshold increases precision, it also dramatically lowers recall, with Figure 3 summarizing the increase in sparsity when increasing the minimum overlap threshold.

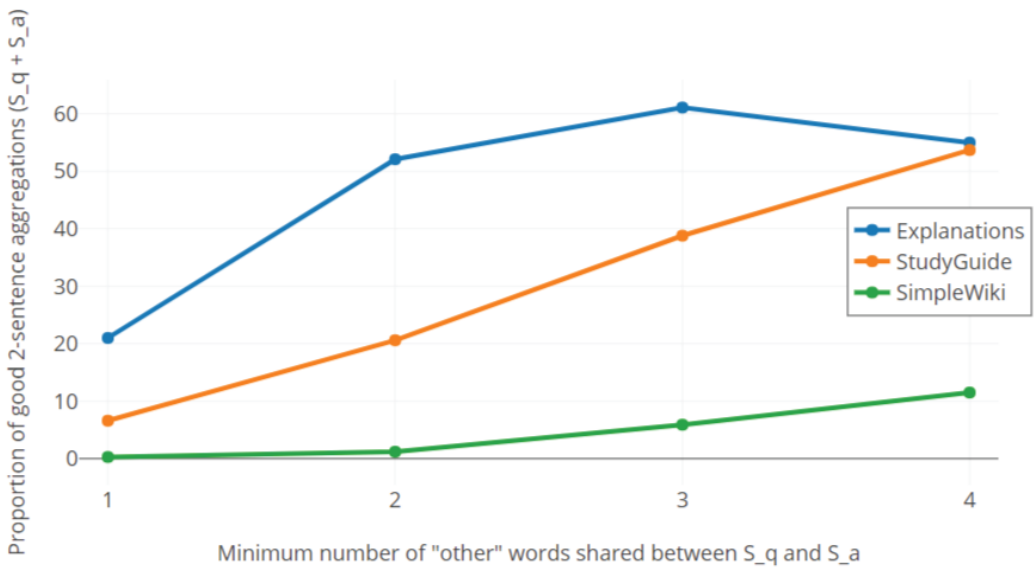

Figure 2: The proportion of "good" 2-sentence aggregations from a given corpus, versus the minimum number of "other" shared words found in both $S_Q$ and $S_A$, but not in the question or answer. "Good" 2-sentence aggregations are defined as the proportion of $Q \leftrightarrow S_Q \leftrightarrow S_A \leftrightarrow A$ aggregations where $S_Q$ and $S_A$ are both rated either *highly* or *possible*, represented by the bolded square in the aggregation quality tables. The graph shows that as the number of "other" shared words increases, so too does the probability of making a good aggregation – thought this often comes at the expense of sparsity.

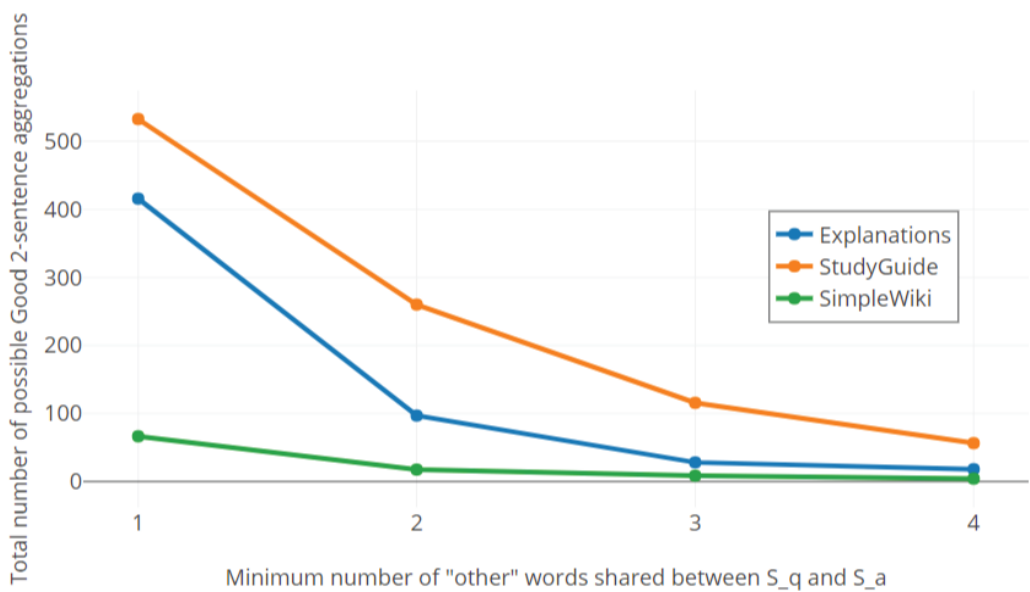

Figure 3: The total number of "good" 2-sentence aggregations from a given corpus, versus the minimum number of "other" shared words found in both $S_Q$ and $S_A$, but not in the question or answer. "Good" 2-sentence aggregations are defined as the proportion of $Q \leftrightarrow S_Q \leftrightarrow S_A \leftrightarrow A$ aggregations where $S_Q$ and $S_A$ are both rated either *highly* or *possible*, represented by the bolded square in the aggregation quality tables. The graph shows that as the number of "other" shared words increases, sparsity begins to dramatically limit the number of good aggregations that can be assembled from a given corpus.

*Explanation Corpus*

*1 or more "other" shared words (1,979 samples)*

|  |  | \multicolumn{4}{c}{$S_A$ Rating} |
|---|---|---|---|---|---|
|  |  | Highly | Possible | Topical | OffTopic |
| $S_Q$ Rating | Highly | **13.4%** | **3.8%** | 3.4% | 1.1% |
|  | Possible | **3.4%** | **0.4%** | 1.1% | 0.6% |
|  | Topical | 7.5% | 1.3% | 8.0% | 4.5% |
|  | Offtopic | 16.3% | 11.0% | 9.1% | 14.9% |

*2 or more "other" shared words (186 samples)*

|  |  | \multicolumn{4}{c}{$S_A$ Rating} |
|---|---|---|---|---|---|
|  |  | Highly | Possible | Topical | OffTopic |
| $S_Q$ Rating | Highly | **44.3%** | – | 0.3% | 0.4% |
|  | Possible | **7.4%** | **0.4%** | 1.1% | 0.4% |
|  | Topical | 4.7% | 7.9% | 16.5% | 0.2% |
|  | Offtopic | 3.9% | 1.2% | 3.1% | 13.1% |

*3 or more "other" shared words (46 samples)*

|  |  | \multicolumn{4}{c}{$S_A$ Rating} |
|---|---|---|---|---|---|
|  |  | Highly | Possible | Topical | OffTopic |
| $S_Q$ Rating | Highly | **53.3%** | – | – | – |
|  | Possible | **4.8%** | **3.0%** | 0.6% | – |
|  | Topical | 4.5% | – | 23.9% | – |
|  | Offtopic | – | – | – | 9.7% |

*4 or more "other" shared words (33 samples)*

|  |  | \multicolumn{4}{c}{$S_A$ Rating} |
|---|---|---|---|---|---|
|  |  | Highly | Possible | Topical | OffTopic |
| $S_Q$ Rating | Highly | **50.0%** | – | – | – |
|  | Possible | **5.0%** | – | – | – |
|  | Topical | – | – | 30.0% | – |
|  | Offtopic | – | – | – | 15.0% |

Table 4: Observed frequencies for aggregating two sentences together with specific utility ratings in the $Q \leftrightarrow S_Q \leftrightarrow S_A \leftrightarrow A$ condition in the COLING Explanations corpus. Here, one sentence in the pair has overlapping terms in the question, the other sentence has overlapping terms in the answer, and both sentences lexically overlap with each other on *one or more, rwo or more, three or more, or four or more* terms that are not found in either the question or answer. Axes represent the individual (nonaggregated) ratings of each sentence (Q or A). The bolded square represents the proportion of lexically connected sentence pairs where utility ratings for both sentences are either *high* or *possible*.

*Study Guide Corpus*

*1 or more "other" shared words (8,069 samples)*

|  |  | \multicolumn{4}{c}{$S_A$ Rating} |
|---|---|---|---|---|---|
|  |  | Highly | Possible | Topical | OffTopic |
| $S_Q$ Rating | Highly | **3.0%** | **0.7%** | 0.7% | 1.3% |
|  | Possible | **2.1%** | **0.8%** | 0.8% | 1.2% |
|  | Topical | 3.8% | 2.0% | 2.7% | 4.4% |
|  | Offtopic | 17.0% | 6.4% | 10.9% | 42.2% |

*2 or more "other" shared words (1,262 samples)*

|  |  | \multicolumn{4}{c}{$S_A$ Rating} |
|---|---|---|---|---|---|
|  |  | Highly | Possible | Topical | OffTopic |
| $S_Q$ Rating | Highly | **12.5%** | **2.9%** | 1.0% | 1.4% |
|  | Possible | **3.6%** | **1.6%** | 1.2% | 0.9% |
|  | Topical | 4.8% | 2.3% | 3.6% | 3.3% |
|  | Offtopic | 5.0% | 8.1% | 14.9% | 33.0% |

*3 or more "other" shared words (298 samples)*

|  |  | \multicolumn{4}{c}{$S_A$ Rating} |
|---|---|---|---|---|---|
|  |  | Highly | Possible | Topical | OffTopic |
| $S_Q$ Rating | Highly | **22.9%** | **5.7%** | 0.8% | 1.8% |
|  | Possible | **6.3%** | **3.9%** | 0.8% | 0.2% |
|  | Topical | 3.9% | 3.3% | 6.3% | 2.4% |
|  | Offtopic | 2.1% | 5.1% | 1.6% | 32.7% |

*4 or more "other" shared words (105 samples)*

|  |  | \multicolumn{4}{c}{$S_A$ Rating} |
|---|---|---|---|---|---|
|  |  | Highly | Possible | Topical | OffTopic |
| $S_Q$ Rating | Highly | **33.7%** | **5.6%** | – | – |
|  | Possible | **6.6%** | **7.8%** | – | – |
|  | Topical | 1.0% | 1.4% | 8.7% | 1.1% |
|  | Offtopic | – | – | 0.8% | 33.4% |

Table 5: Observed frequencies for aggregating two sentences together with specific utility ratings in the $Q \leftrightarrow S_Q \leftrightarrow S_A \leftrightarrow A$ condition in the Study Guide corpus. Here, one sentence in the pair has overlapping terms in the question, the other sentence has overlapping terms in the answer, and both sentences lexically overlap with each other on *one or more, rwo or more, three or more, or four or more* terms that are not found in either the question or answer. Axes represent the individual (nonaggregated) ratings of each sentence (Q or A). The bolded square represents the proportion of lexically connected sentence pairs where utility ratings for both sentences are either *high* or *possible*.

*Simple Wikipedia Corpus*

*1 or more "other" shared words (23,750 samples)*

|  |  | $S_A$ Rating | | | |
|---|---|---|---|---|---|
|  |  | *Highly* | *Possible* | *Topical* | *OffTopic* |
| $S_Q$ Rating | *Highly* | **0.04%** | **0.04%** | 0.06% | 0.4% |
|  | *Possible* | **0.1%** | **0.1%** | 0.3% | 1.8% |
|  | *Topical* | 1.3% | 2.3% | 2.4% | 2.1% |
|  | *Offtopic* | 2.4% | 2.0% | 7.2% | 82.9% |

*2 or more "other" shared words (1,464 samples)*

|  |  | $S_A$ Rating | | | |
|---|---|---|---|---|---|
|  |  | *Highly* | *Possible* | *Topical* | *OffTopic* |
| $S_Q$ Rating | *Highly* | **0.3%** | – | 0.08% | 0.3% |
|  | *Possible* | **0.4%** | **0.5%** | 1.0% | 5.1% |
|  | *Topical* | 0.1% | 0.4% | 0.7% | 2.5% |
|  | *Offtopic* | 3.1% | 2.0% | 5.3% | 78.3% |

*3 or more "other" shared words (146 samples)*

|  |  | $S_A$ Rating | | | |
|---|---|---|---|---|---|
|  |  | *Highly* | *Possible* | *Topical* | *OffTopic* |
| $S_Q$ Rating | *Highly* | **2.6%** | – | – | – |
|  | *Possible* | **0.6%** | **2.7%** | 0.2% | 9.2% |
|  | *Topical* | 0.2% | – | 1.6% | 3.1% |
|  | *Offtopic* | 2.4% | 0.8% | 3.6% | 73.0% |

*4 or more "other" shared words (37 samples)*

|  |  | $S_A$ Rating | | | |
|---|---|---|---|---|---|
|  |  | *Highly* | *Possible* | *Topical* | *OffTopic* |
| $S_Q$ Rating | *Highly* | **6.3%** | – | – | – |
|  | *Possible* | – | **5.2%** | – | 4.4% |
|  | *Topical* | – | – | 4.0% | 2.4% |
|  | *Offtopic* | – | – | – | 77.8% |

Table 6: Observed frequencies for aggregating two sentences together with specific utility ratings in the $Q \leftrightarrow S_Q \leftrightarrow S_A \leftrightarrow A$ condition in the Simple Wikipedia corpus. Here, one sentence in the pair has overlapping terms in the question, the other sentence has overlapping terms in the answer, and both sentences lexically overlap with each other on *one or more, rwo or more, three or more, or four or more* terms that are not found in either the question or answer. Axes represent the individual (nonaggregated) ratings of each sentence (Q or A). The bolded square represents the proportion of lexically connected sentence pairs where utility ratings for both sentences are either *high* or *possible*.